\documentclass[conference]{IEEEtran}
\IEEEoverridecommandlockouts
\usepackage{cite}
\usepackage{amsmath,amssymb,amsfonts}
\usepackage{algorithm}
\usepackage{algorithmicx}
\usepackage{algpseudocode} 
\usepackage{graphicx}
\usepackage{textcomp}
\usepackage{xcolor}
\usepackage{booktabs}
\usepackage{multirow}
\usepackage{subfigure}
\usepackage{soul}
\usepackage{url}
\usepackage[colorlinks,
            linkcolor=red,
            anchorcolor=blue,
            citecolor=green
            ]{hyperref}
\def\BibTeX{{\rm B\kern-.05em{\sc i\kern-.025em b}\kern-.08em
    T\kern-.1667em\lower.7ex\hbox{E}\kern-.125emX}}
\begin{document}

\newcommand{\notice}[1]{\textcolor{blue}{#1}}
\setlength{\textfloatsep}{1pt plus 0.2pt minus 0.1pt}
\setlength{\intextsep}{1pt plus 0.2pt minus 0.1pt}

\def\arraystretch{0.60}

\title{Boosting the Adversarial Transferability of Surrogate Models with Dark Knowledge}

\author{\IEEEauthorblockN{Dingcheng Yang$^{1,2}$, Zihao Xiao$^2$, Wenjian Yu$^{1,*}$\thanks{$^*$corresponding author.}}
\IEEEauthorblockA{$^1$Dept. Computer Science \& Tech., BNRist, Tsinghua University, Beijing, China.  \\
$^2$RealAI. \\
ydc19@mails.tsinghua.edu.cn, zihao.xiao@realai.ai, yu-wj@tsinghua.edu.cn
}
}

\maketitle

\begin{abstract}
Deep neural networks (DNNs) are vulnerable to adversarial examples. And, the adversarial examples have transferability, which means that an adversarial example for a DNN model can fool another model with a non-trivial probability. This gave birth to the transfer-based attack where the adversarial examples generated by a surrogate model are used to conduct black-box attacks. There are some work on generating the adversarial examples from a given surrogate model with better transferability. However, training a special surrogate model to generate adversarial examples with better transferability is relatively under-explored. This paper proposes a method for training a surrogate model with dark knowledge to boost the transferability of the adversarial examples generated by the surrogate model. This trained surrogate model is named dark surrogate model (DSM). The proposed method for training a DSM consists of two key components: a teacher model extracting dark knowledge, and the mixing augmentation skill enhancing dark knowledge of training data. We conducted extensive experiments to show that the proposed method can substantially improve the adversarial transferability of surrogate models across different architectures of surrogate models and optimizers for generating adversarial examples, and it can be applied to other scenarios of transfer-based attack that contain dark knowledge, like face verification. Our code is publicly available at \url{https://github.com/ydc123/Dark_Surrogate_Model}.
\end{abstract}


\section{Introduction}
Deep neural networks (DNNs)  have achieved substantial success on many computer vision tasks. However, they are shown to be vulnerable to adversarial examples. Adversarial examples are carefully crafted data which could fool the DNNs by adding imperceptible noise on legitimate data. The generation of adversarial examples have been extensively researched in recent years. This is primarily due to its potential for preventing the malicious use of DNNs~\cite{shi2021adversarial} and serving as a reliable evaluation criterion for DNN security~\cite{zhao2022towards}, particularly in safety-critical scenarios like face verification.

The transferability of adversarial examples has attracted much attention. It means that, an adversarial example that fools one DNN model can fool another DNN model with a non-trivial probability. Thus, an adversary can train a surrogate model locally (training stage), and then generate adversarial examples to fool the surrogate model (generating stage). Finally, the generated adversarial examples can be directly used to attack an unknown black-box victim model (attacking stage). This process is called \emph{transfer-based adversarial attack}.

    The technique of adversarial example optimizer has been proposed for generating highly transferable adversarial examples~\cite{dong2018boosting,xie2019improving} (in generating stage). In contrast, we aim to train a better surrogate model (in training stage) so that it could yield adversarial examples with better success rates of transfer-based attacks. In analogy to the commonly used term ``the transferability of adversarial example'', we propose the concept ``the adversarial transferability of surrogate models'' to describe the ability of surrogate models on generating better adversarial examples for transfer-based attacks, using a fixed adversarial example optimizer. There are just a few works trying to train a surrogate model with better adversarial transferability \cite{cui2020substitute,springer2021a}. Specifically, adversarial training was used in~\cite{springer2021a} to improve the transferability, but at a significant computational cost. In \cite{cui2020substitute}, the knowledge distillation~\cite{hinton2015distilling} was applied to improve the transferability without incurring excessive time overhead. However, it relies on a scheme of ensemble attack with multiple surrogate models and a combination of soft labels and one-hot labels.

Labels and data are two important components in training DNNs. Although the one-hot label is extensively used in normal DNN training, we notice that it does not well describe a data, because an image often contains the features of similar classes and even multiple objects in addition to the features of the true class.
In contrast to one-hot labels, the soft labels which are the predicted distributions from a teacher model contain abundant information of image data, and have been used in 
knowledge distillation~\cite{hinton2015distilling} 
for compressing neural networks. The soft label is also known as ``dark knowledge''~\cite{hinton2015distilling}. In this work, we propose that the dark knowledge is the key recipe to boost the adversarial transferability of surrogate models. Therefore, we propose to use the soft label to train the surrogate model, and enhance the dark knowledge by applying mixing augmentation skills to training data~\cite{devries2017cutout,zhang2018mixup,yun2019cutmix}.

The surrogate model trained with dark knowledge is called  ``dark'' surrogate model (DSM) in this work. The proposed method modifies the training stage, which enhances the dark knowledge by applying mixing augmentation to the training data and using soft labels extracted from a pretrained teacher model. We have conducted extensive experiments on attacking image classification models
to show that the proposed method remarkably and consistently improves the adversarial transferability of surrogate models.
In addition, the proposed method can be applied to other transfer-based attack scenarios that contain dark knowledge, such as face verification, image retrieval, and text classification, to improve the success rate of the transfer-based attack. As an example, the experiments on applying DSM to attack face verification models are presented.

The major contributions and results are as follows.
\begin{itemize}
        \item For improving the success rates of the transfer-based adversarial attack, we propose to use the dark knowledge during the training of the surrogate model, so as to obtain a ``dark'' surrogate model (DSM). The method for training the DSM is proposed, which modifies two key components of DNN training: labels and data, to make full usage of dark knowledge. Firstly, a pretrained DNN model, regarded as a teacher model, is employed to generate soft labels with dark knowledge. Secondly, the mixing augmentation skills are applied to enhance the dark knowledge of the training data explicitly.
    \item Extensive experiments on image classification are conducted to validate the proposed method. At first, the DSM is trained by using a pretrained model of the same architecture as the teacher model. Compared to directly using the pretrained model as the surrogate model, the proposed method improves the success rates of the untargeted attack 
    by a TI-DI-MI-TG optimizer~\cite{he2022boosting} on nine victim models by up to \textbf{21.6\%}, \textbf{23.6\%} and \textbf{11.0\%} for the ResNet18, DenseNet121 and MobileNetv2 based surrogate models, respectively. Then, by using different teacher models, the maximum increments of attack success rate can be further improved to \textbf{25.7\%}, \textbf{36.8\%} and \textbf{26.3\%}, respectively. Experimental results also validate that the proposed method performs better than the related work based on knowledge distillation~\cite{cui2020substitute}.
        \item We have also applied the proposed method to the problem of attacking face verification models. On the widely-used ArcFace model~\cite{deng2019arcface}, the proposed method improves the success rates of dodging attack by $\textbf{12.9\%}$ and impersonate attack by $\textbf{16.2\%}$.
\end{itemize}

\section{Related works}
A DNN model for classification can be considered as a function $f(x; \theta): \mathbb{R}^d \rightarrow \mathbb{R}^K$, where $K$ is the number of classes, $\theta$ denotes all the parameters, $x \in \mathbb{R}^d$ is an image, $d$ denotes the dimensionality of $x$, and the predicted label is $\text{argmax}_{1\le i \le K} \, f(x; \theta)_i$.

Given an image $x$ and its corresponding label $y$, an untargeted adversarial example (the example which is misclassified) can be generated to fool a DNN model parameterized by $\theta$ through maximizing a cross-entropy loss function:
\begin{equation}
    \begin{split}
        & x^* \!=\! \text{argmax}_{x'} \mathbf{CE}(e_y, \mathbb{S}(f(x';\theta)))~, \text{s.t.}~ \|x'-x\| \le \epsilon~,
        \label{eq:eq_un}
    \end{split}
\end{equation}
where $e_y$ denotes a one-hot vector with true label $y$, and the cross-entropy loss function $\mathbf{CE}(p,q)$ is defined as $\mathbf{CE}(p,q)=-\sum_i p_i \log q_i$. The softmax function $\mathbb{S}: \mathbb{R}^K \rightarrow \mathbb{R}^K$ is used to normalize the outputs of a DNN to a probability distribution, which means $\mathbb{S}(z)_i=\exp(z_i)/\sum_{j=1}^K \exp(z_j)$. The $\| \cdot \|$ denotes a norm function, and we focus on $\mathcal{L}_{\infty}$ norm in this paper. The $\epsilon$ is the maximum allowed magnitude of perturbation. The generated adversarial example $x^*$ looks similarly to $x$ but can fool the DNN model parameterized by $\theta$ (also called victim model).

However, the victim model is often inaccessible in practice. To attack a black-box victim model, we should first train a white-box surrogate model $\theta$, and use it to generate the adversarial example $x^*$. This adversarial example is then directly used to attack the victim model. Normally, a surrogate model $\theta$ is trained by solving the following optimization problem with a stochastic
gradient descent optimizer:
\begin{equation}
    \begin{aligned}
        \theta = \text{argmin}_{\theta'} \mathbf{CE}(e_y, \mathbb{S}(f(x;\theta'))) ~.
    \end{aligned}
    \label{eq:eq1}
\end{equation}

Many optimizers were proposed for generating untargeted adversarial examples by solving~\eqref{eq:eq_un}, using a surrogate model $\theta$ that was trained by solving~\eqref{eq:eq1}, such as the one-step method FGSM~\cite{goodfellow2014explaining}, the MI-FGSM~\cite{dong2018boosting} that utilizes a momentum factor $\mu$, the diverse inputs method (DIM)~\cite{xie2019improving} that augments the inputs with a pre-defined probability $p_t$, and the translation-invariant method (TIM)~\cite{dong2019evading} that convolves the gradient with a pre-defined kernel $W$. Recently, a more powerful optimizer has been proposed that utilizes transformed gradient (TG) to generate adversarial examples, and its combination with the above methods leads to a more effective TI-DI-MI-TG method~\cite{he2022boosting}. These methods can be easily extended for the targeted attack, i.e. generating an adversarial example $x^*$ which is misclassified as a pre-defined target label $y_t$. It has been recently shown that targeted attacks can be boosted by a logits-based loss and running more iterations~\cite{zhao2021on}.

Unlike the rapidly developing adversarial example optimizers, only a few of works were devoted to training a better surrogate model for the transfer-based attack. They include the knowledge-distillation based method~\cite{cui2020substitute} and the recent work~\cite{springer2021a} showing that a slightly robust model has better adversarial transferability. Notice that, the method in~\cite{springer2021a} costs large computational time for training
the slightly robust model.

In the knowledge-distillation based method~\cite{cui2020substitute}, a surrogate model is distilled using multiple teacher models. This method is inspired by previous works on ensemble attack~\cite{liu2016delving}, which show that attacking multiple surrogate models simultaneously is more effective than attacking a single surrogate model. By distilling from multiple teacher models, the resulting surrogate model shares similar characteristics with all the teachers, thereby mimicking the ensemble attack. Specifically, suppose there are $M$ teacher models $F_1,\cdots,F_M$, a surrogate model $\theta$ is trained by solving the following optimization problem:
\begin{equation}
\begin{split}
    & \theta = \text{argmin}_{\theta'} \mathbf{CE}(\tilde{y}, \mathbb{S}(f(x;\theta')))~,\\
    & \text{where}\quad \tilde{y}= \frac{\beta_{KD}}{M}\sum_{i=1}^M \mathbb{S}(F_i(x))+(1-\beta_{KD}) e_y ~.
    \label{eq:eq_KD}
\end{split}
\end{equation}

 The $F_i(x)$ is the output of the $i$-th teacher for image $x$, $\tilde{y}$ is a soft label generated for training the surrogate model, and $\beta_{KD}$ is a hyper-parameter.

There are several studies on data augmentation by mixing multiple data for image classification, including Cutout~\cite{devries2017cutout}, Mixup~\cite{zhang2018mixup} and CutMix~\cite{yun2019cutmix}. In Section IV.B, we will show that after enriching the dark knowledge of the images by these skills, the adversarial transferability can be further improved when using a teacher model to extract dark knowledge from the augmented images. To the best of our knowledge, we are the first to use conventional data augmentations in the \emph{training} stage to boost transfer-based attacks, as opposed prior works that augment data during the \emph{generating} stage~\cite{xie2019improving,dong2019evading,he2022boosting}.
\section{Methodology}

\begin{figure}[b]
  \setlength{\abovecaptionskip}{0 cm}
  \centering
    \includegraphics[width=3.5in]{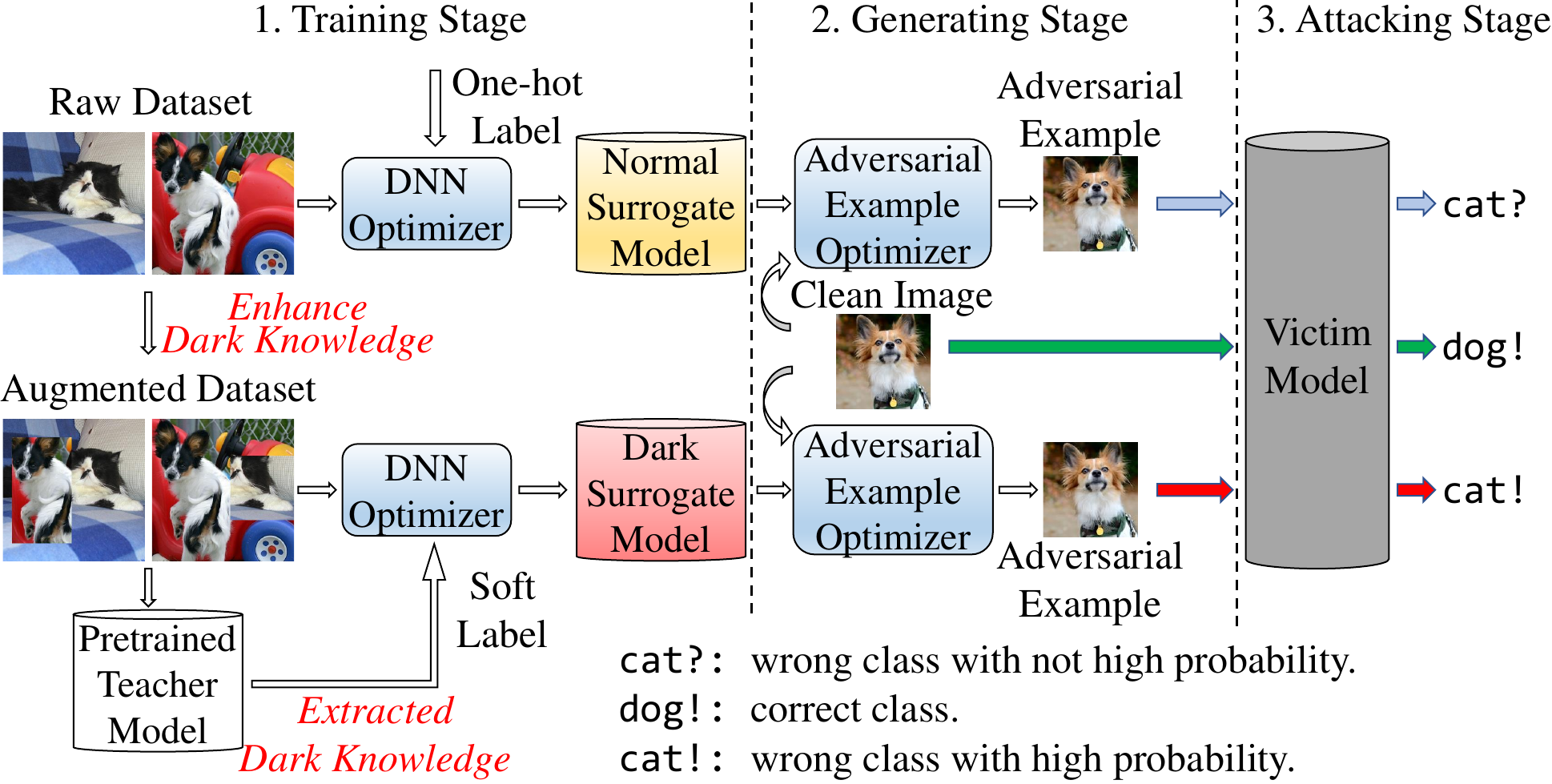}
  \caption{An illustration of previous work and the proposed method for generating adversarial examples.
  }
    \label{fig:pipeline}
\end{figure}

In this section, we propose the dark surrogate model (DSM) to yield adversarial examples with better transferability, which is illustrated in Fig.~\ref{fig:pipeline}. We first introduce our idea of refining labels with dark knowledge. Then, we apply mixing augmentations to enhance the dark knowledge of training data. Finally, we describe the algorithm for training the proposed DSM.
\subsection{Refining Labels Using Dark Knowledge}

Given an image $x$ and its label $y$, the optimization problem~\eqref{eq:eq1} converges a minimum value only if the predicted distribution $\mathbb{S}(f(x;\theta))$ equals the one-hot label $e_y$, which means $f(x;\theta)_y-\max_{i\neq y}f(x;\theta)_i=\infty$, indicating that the trained surrogate model output an extremely high confidence score for the true class. However, the fitting target $e_y$ does not describe an image well because an image often contains features of similar classes. For example, ImageNet~\cite{deng2009imagenet}, the most famous dataset for image classification, is a hierarchical dataset that contains many subcategories belonging to the category ``dog'', such as ``papillon'', ``chihuahua'', ``Maltese dog''. An image of ``papillon'' will have the features of other ``dog'' categories. Moreover, there may even be multiple objects appearing in an image. An example is illustrated in Fig.~\ref{fig:pipeline}, where there are two images in the Raw Dataset labeled as ``Persian cat'' and ``papillon'', respectively, while they possess features of other cats, dogs, as well as pillows and cars. Even if the model achieves high accuracy on classification, the model trained with one-hot labels can not fully extract the features of an image for every class. This will be harmful for adversarial transferability of surrogate model, which directly depends on the working mechanism of the trained surrogate model, i.e. how it thinks ``an image looks like a dog instead of a goldfish''.

To overcome this weakness, we propose to leverage a pretrained teacher model to extract the dark knowledge from the training data, which is then utilized to train the surrogate model. Specifically, the predicted probability distribution of the teacher model serves as a soft label, which provides more information compared to the one-hot label, such as ``which 2’s look like 3’s and which look like 7’s''~\cite{hinton2015distilling}. This information can help the surrogate model to learn image features better, and thus yield more transferable adversarial examples. Given a pretrained teacher model parameterized by $\theta_0$, we can train a dark surrogate model parameterized by $\theta_d$ through solving the following optimization problem:
\begin{align}
\label{eq:eq_dark}
    \theta_d &= \text{argmin}_{\theta}\, \, \mathbf{CE}(\mathbb{S}(f(x;\theta_0)), \mathbb{S}(f(x;\theta))) ~.
\end{align}

The major difference to the normal training~\eqref{eq:eq1} is that the dark knowledge $\mathbb{S}(f(x;\theta_0))$ produced by the teacher model is used as the label.

Our work shares a similar process with the previous work~\cite{cui2020substitute}, but we are motivated by a different goal and complement their work. 
 Specifically, the objective in~\cite{cui2020substitute} is to make the surrogate model similar to multiple teacher models, and the success of~\cite{cui2020substitute} is entirely attributed to the ensemble attack~\cite{liu2016delving}, without any discussion on dark knowledge. This makes the method in \cite{cui2020substitute} incomplete and lack theoretic support for the case of using a single teacher model. However, experimental results to be presented in Section IV show that using only a single teacher model can also improves the adversarial transferability. This improvement can not be explained by~\cite{cui2020substitute}, but we provide a theoretical interpretation for this phenomenon based on the concept of dark knowledge. Additionally, while one-hot labels ($e_y$ term in~\eqref{eq:eq_KD}) are still employed in~\cite{cui2020substitute}, our perspective based on dark knowledge indicates that they are harmful for adversarial transferability and thus we do not use them. In Section IV.C, we will conduct a comparison experiment with~\cite{cui2020substitute} to highlight the differences between the two works.

\subsection{Enhancing Dark Knowledge of Training Data}

\begin{figure}[h]
  \setlength{\abovecaptionskip}{0 cm}
  \centering
    \includegraphics[width=2.0in]{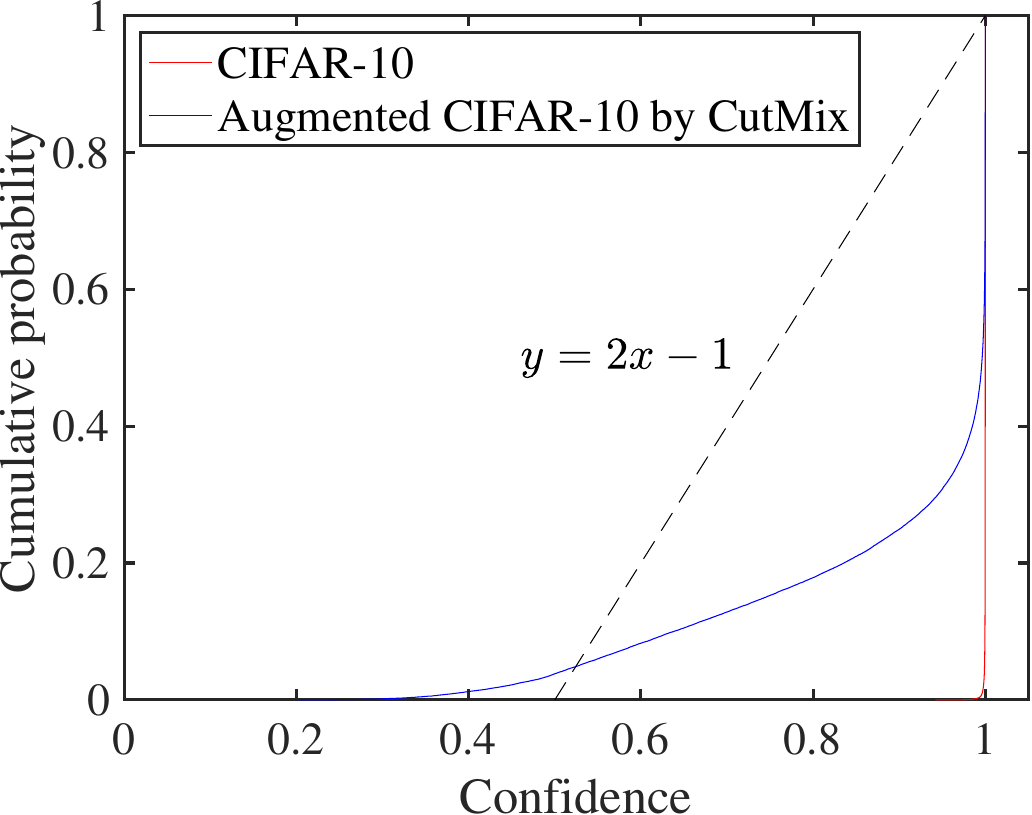}
  \caption{The empirical cumulative distribution function (CDF) of confidence of soft labels generated by an ResNet18 teacher model.}
    \label{fig:pred}
\end{figure}

Although the soft label in~\eqref{eq:eq_dark} involves dark knowledge and thus is better than the one-hot label, it is still close to the one-hot label since the teacher model is obtained by training with the one-hot labels. To illustrate this point, we denote the confidence of a soft label $\tilde{y}$ as $\max_{i=1}^K \tilde{y}_i$. Thus, the confidence of a one-hot label achieves the maximum value of 1. Then, we train a ResNet18 teacher model on CIFAR-10 and provide a visualization of the empirical cumulative distribution function (CDF) of the confidences of soft labels it generates on the training images of CIFAR-10, as shown in Fig.~\ref{fig:pred}. Fig.~\ref{fig:pred} shows that the empirical CDF of confidences on CIFAR-10 (red curve) is similar to the CDF of one-hot labels, namely a straight line with $x=1$, which greatly weakens the effect of dark knowledge on boosting adversarial transferability.

To overcome this weakness, we propose to enhance the dark knowledge of training data by leveraging the data augmentation skills which explicitly mix a pair of images to synthesize image data containing features of different classes. Given an original image $x$, we consider three popular skills of mixing augmentations in this work:
\begin{itemize}
    \item Cutout~\cite{devries2017cutout}, which randomly masks a fixed-size area of $x$ to zero. The size of mask is set to 112$\times$112 in this work.
    \item Mixup~\cite{zhang2018mixup}, which randomly samples a reference image $x'$ and make a combination with $x$. This generates an image $\tilde{x}=\lambda x+(1-\lambda) x'$, where $\lambda \sim U(0,1)$ in this work. For this data $\tilde{x}$, a soft label $\tilde{y}=\lambda e_y+(1-\lambda) e_{y'}$ should be used during the training, where $y$ and $y'$ are the true classes of $x$ and $x'$, respectively.
    \item CutMix~\cite{yun2019cutmix}, which randomly copies a rectangle area of $x'$ to paste into $x$.
  If the area ratio of the rectangle to the whole image is $1-\lambda$ where $\lambda \sim U(0,1)$ in this work, a soft label $\tilde{y}=\lambda e_y+(1-\lambda) e_{y'}$ is used for training, and $y$ and $y'$ are the true classes of $x$ and $x'$, respectively.
\end{itemize}

\begin{figure}
  \setlength{\abovecaptionskip}{0 cm}
  \centering
    \includegraphics[width=3.3in]{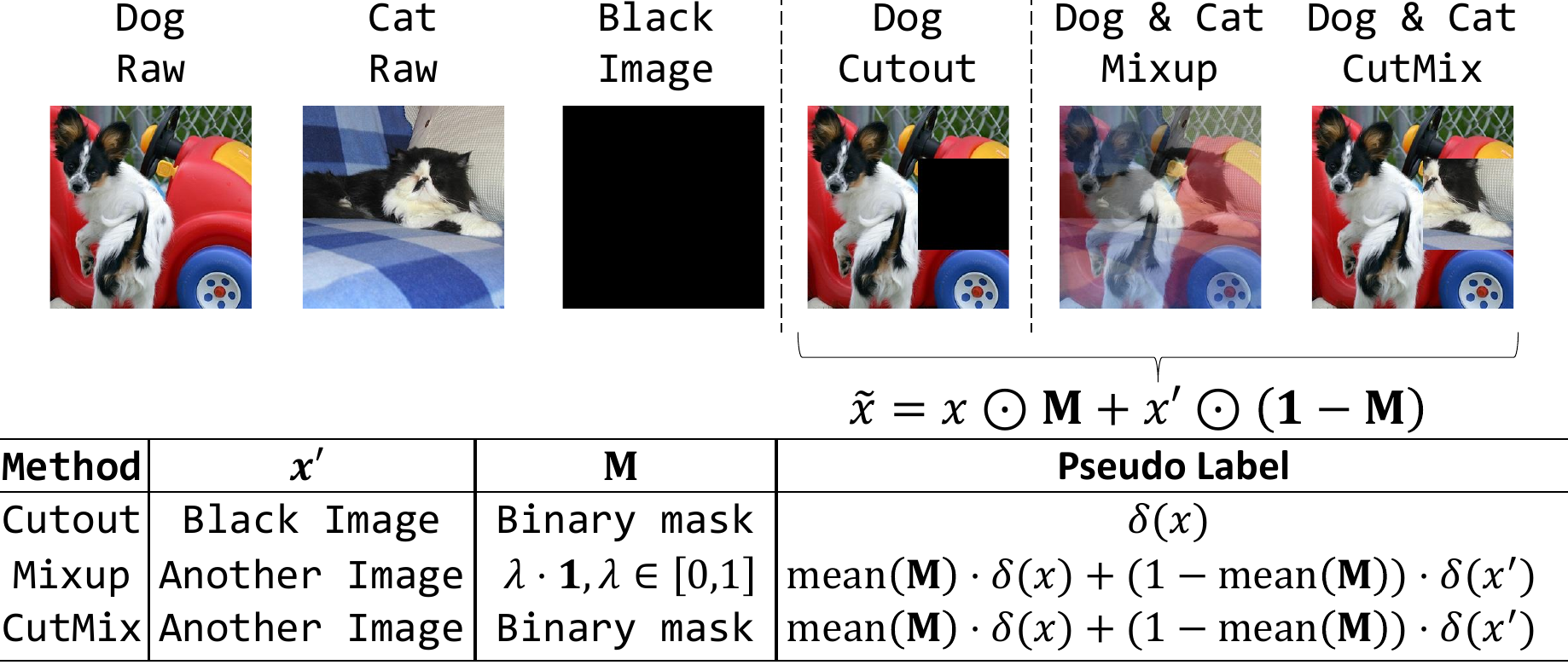}
  \caption{
  An illustration of different data augmentation skills. The details of these skills are explained at bottom.  $\delta(x)$ denotes the one-hot label of $x$.
  }
    \label{fig:demo}
\end{figure}

The data generated with the three mixing augmentation skills can be unified as $\tilde{x}=x \odot \mathbf{M} + x' \odot (\mathbf{1} - \mathbf{M})$, where $\mathbf{M}$ is a tensor of the same shape as $x$, $\odot$ is an element-wise product, and $\mathbf{1}$ is an all-one tensor. Fig.~\ref{fig:demo} illustrates these skills.

 To demonstrate the effectiveness of mixing augmentation in mitigating the issue of soft labels being too similar to one-hot labels, we take CutMix as an example and apply it to the training images of CIFAR-10 to generate 50000 augmented images. We then plot the CDF of the confidences of generated soft labels on these augmented images in the blue curve of Fig.~\ref{fig:pred}, which exhibit clear deviation from the red curve over about 40\% data points, demonstrating that the mixing augmentations can construct data with more dark knowledge.
 
 Moreover, Fig.~\ref{fig:pred} also reveals a huge discrepancy between the CDF of confidence of soft labels generated by the ResNet18 teacher model and those for the standard CutMix skill. Specifically, the soft labels employed in the standard CutMix is $\tilde{y}=\lambda e_y + (1-\lambda)e_{y'}$ for an augmented image obtained by mixing by two images that belong to the category of $y$ and $y'$, respectively, where $\lambda \sim U(0, 1)$. As a result, the confidence of the soft label is $\max(\lambda, 1-\lambda)$ and should obey a uniform distribution on $[0.5, 1]$, with a CDF of $y=2x-1$, which is completely different from the trend of the blue curve depicted in Fig.~\ref{fig:pred}. This difference raises the question of whether the heuristic labeling strategy employed in CutMix is a reasonable strategy for improving adversarial transferability. Although CutMix has been shown to be effective on image classification, as discussed in Section III.A, higher classification accuracy does not necessarily correspond to better adversarial transferability. Notably, the labeling strategy of CutMix skill has a similar limitation to one-hot labels as it only considers the features of the categories to which the mixed images belong, while ignoring those of most other categories. This limitation is not aligned with the motivation of this work and can also be observed in other mixing augmentation skills. Therefore, CutMix and other mixing augmentation skills used in our method are only to enhance the dark knowledge of training data. The soft labels are obtained through the method introduced in Section III.A, specifically, by using the predicted probability distribution of the teacher model for the augmented images. This step is also illustrated in Fig.~\ref{fig:pipeline} and will be described as Algorithm 1 in Section III.C.

The experimental results in Section IV.B will show that the adversarial transferability of DSMs can be further improved by enhancing the dark knowledge of the training data with these mixing augmentations. Furthermore, Section IV.B also shows that the standard mixing augmentation will impair the adversarial transferability of surrogate models. It is noteworthy that the only difference between training with the standard mixing augmentation skills and the proposed DSM is the presence or absence of dark knowledge in the labeling strategy. This finding further emphasizes the crucial role of dark knowledge in boosting adversarial transferability.
\subsection{The Proposed Algorithm for Training DSM}
Combining the ideas in last two subsections, we propose the approach of training the DSM to boost the adversarial transferability of surrogate models, described as Algorithm 1. At each iteration during the training, we first enhance the dark knowledge by the mixing augmentation skills (Step 6), then train the surrogate model with dark knowledge extracted by the teacher model described with parameters $\theta_0$ (Step 7).

\begin{algorithm}[h!]
\caption{Training the DSM for transfer-based attack}\label{alg:DSM}
\begin{flushleft}
\textbf{Input:} Batch size $m$, learning rate $\eta$, training dataset $\mathcal{D_T}$, a  (pretrained) teacher model with parameters $\theta_0$, dark surrogate DNN model parameterized by $\theta_d$.\\
\textbf{Output:} the dark surrogate model with parameters $\theta_d$.\\
\end{flushleft}
\begin{algorithmic}[1]

\State Randomly initialize the parameters $\theta_d$. 
\Repeat \Comment{solve the optimization problem (8)}
    \State Read mini-batch $\{x_1,\cdots,x_m\}$ from $\mathcal{D_T}$.
\State $L \gets 0$.
    \For{$i \gets 1$ to $m$}
        \State Apply the mixing augmentation on $x_i$ to obtain an augmented image $x^{mix}_i$. 
        \State $L \gets L +  \mathbf{CE}(\mathbb{S}(f(x^{mix}_i;\theta_0)),\mathbb{S}(f(x^{mix}_i;\theta_d)))$
    \EndFor
    \State $\theta_d \gets \theta_d - \eta \nabla_{\theta_d} L$
\Until{parameters $\theta_d$ are converged}

\end{algorithmic}
\end{algorithm}

Notice that any pretrained model for the same classification problem can be used as the teacher model. A simple choice of teacher model is the one with the same architecture as the DSM $\theta_d$ and trained by solving problem (2). Section IV.B will show that a teacher model with a different architecture from the DSM is also useful and sometimes makes the DSM exhibit better adversarial transferability of surrogate models. In addition, the proposed approach can be naturally combined with prior work on improving the adversarial transferability of surrogate models, through using their released model as a teacher model, as shown in Section IV.D.

Finally, the proposed approach can be applied to other scenarios of transfer-based attack that contain dark knowledge, like face verification. Training a face verification model consists of two steps, i.e., training a facial classifier and obtaining an embedding model based on that classifier. An adversary can train a facial classifier based on Algorithm 1 to obtain an embedding model. The obtained embedding model can be used as a surrogate to attack a black-box face verification model. We will show that the facial classifier trained by the proposed approach yields an embedding model with better adversarial transferability, with the experiments presented in Section IV.E.
\section{Experimental Results}

In this section, we demonstrate the effectiveness of the proposed dark surrogate model (DSM) with experiments. We begin by evaluating the effectiveness of DSM on attacking image classification models. The ResNet18 (RN18)~\cite{he2016deep}, DenseNet121 (DN121)~\cite{huang2017densely} and MobileNetv2 (MNv2)~\cite{sandler2018mobilenetv2} are chosen as the surrogate models. Unless explicitly stated, the CutMix skill is used for training DSM, and the teacher model employed is a normal pretrained model (trained with one-hot labels) with the same architecture as the DSM. The first three subsections are dedicated to the untargeted attack of image classifiaction models, including the comparison with the related work~\cite{cui2020substitute}. Then, the results of targeted attack are presented, which involves the combination of the proposed DSM and the method of training slightly robust model~\cite{springer2021a}. Lastly, the results on attacking face verification models are presented.

Adversarial examples are crafted with a maximum perturbation of $\epsilon=16$ unless explicitly stated. We consider three adversarial example optimizers: FGSM~\cite{goodfellow2014explaining}, MI-FGSM~\cite{dong2018boosting}, and TI-DI-MI-TG~\cite{he2022boosting}. If the optimizer is not explicitly stated, the strongest TI-DI-MI-TG optimizer is employed. 
For the hyper-parameters of adversarial example optimizer, we set the step size $\beta$ to 2, the momentum factor $\mu$ to 1.0, the probability of transformation $p_t$ to 1.0, the size of kernel $W$ to $7 \times 7$, the number of iterations $N$ for untargeted attack to 10, following~\cite{he2022boosting}. For targeted attack  we set $N$ to 200 and optimize the logits-based loss following the suggestion in~\cite{zhao2021on}.

Two image classification datasets are considered, including the small CIFAR-10 dataset and the large ImageNet dataset. In the experiments of CIFAR-10, the models are trained 200 epochs. We set the batch size to 256, the weight decay to $10^{-4}$. The learning rate is set to 0.1 and updated by a cosine annealing scheduler. In the experiments of ImageNet, we follow the PyTorch official example\footnote{https://github.com/pytorch/examples/tree/master/imagenet} to train the models. We randomly sample 1000 images from CIFAR-10 dataset for generating adversarial examples, while for the experiments of ImageNet, the adversarial examples are generated on ImageNet-compatible dataset\footnote{https://www.kaggle.com/google-brain/nips-2017-adversarial-learning-development-set} since it was widely used in previous works~\cite{dong2018boosting}. This dataset comprises 1000 images and provides a true label and a target label of each image for untargeted and targeted attack, respectively.

\subsection{Detailed Results of the Proposed Method}

We first preliminarily show the performance of DSM on a simple dataset, CIFAR-10. Specifically, we train three normal surrogate models, ResNet18 (RN18), DenseNet121 (DN121) and MobileNetV2 (MNV2), on CIFAR-10, and then employ them as the teacher models to train DSMs with the same architecture. We denote a dark RN18 model trained without mixing augmentation as DSM(RN18, None), and a dark RN18 model trained with CutMix as DSM(RN18, CutMix). When there is no ambiguity, DSM(RN18, CutMix) is abbreviated as DSM(RN18), as CutMix is the default augmentation technique used in this paper. The naming convention of other models is done similarly. We employ the FGSM optimizer to efficiently generate adversarial examples by these normal/dark surrogate models, and use these normally trained models as victim models to evaluate the adversarial transferability. We list the experimental results in Table~\ref{tab:cifar}, which first show that the dark surrogate models are only slightly better than or comparable to normal surrogate models when no mixing augmentation is used. This is because the dark knowledge extracted by the teacher models is too similar to the one-hot labels, as shown in Fig.~\ref{fig:pred}. Then, Table~\ref{tab:cifar} demonstrates that the attack success rates are remarkably improved by up to \textbf{27.0\%} when CutMix is used to enhance the dark knowledge of the training data, emphasizing the crucial role of mixing augmentation.
\begin{table}[h]
    \centering
      \setlength{\abovecaptionskip}{0 cm}
    \caption{
    {The success rates (\%) of untargeted attacks on CIFAR-10 dataset. $-$ indicates the white-box attack.}}
    \label{tab:cifar}
    \begin{tabular}{@{}c@{~}c@{~}c@{~}c@{~}c@{}}
    \toprule
         Surrogate model& RN18 & DN121 & MNv2 \\
         \cmidrule(r){1-1}
         \cmidrule(r){2-4}
         Normal RN18 & - & 73.0 & 70.0\\
         DSM(RN18,None) & - & 74.1 & 68.2\\
         DSM(RN18,CutMix) & - & \textbf{90.7} & \textbf{87.8}\\
         \midrule
         Normal DN121 & 45.2 & - & 65.1 \\
         DSM(DN121,None) & 46.2 & - & 65.5\\
         DSM(DN121,CutMix) & \textbf{72.2} & - & \textbf{88.4}\\
         \midrule
         Normal MNv2 & 38.2 & 60.6 & -\\
         DSM(MNv2,None) & 41.3 & 59.7 & -\\
         DSM(MNv2,CutMix) & \textbf{64.2} & \textbf{85.7} & -\\
         \bottomrule
    \end{tabular}
\end{table}
\begin{table*}[t]
  \setlength{\abovecaptionskip}{0 cm}
    \centering
    \caption{The success rates (\%) of untargeted attacks on ImageNet-compatible dataset with different optimizers.}
    \label{tab:imagenet-un}

    \begin{tabular}{@{}c@{~}c@{~}c@{~}c@{~}c@{~}c@{~}c@{~}c@{~}c@{~}c@{~}c@{}}
        \toprule
         Optimizer & Surrogate model & Inc-v3 & Inc-v4 & IncRes-v2 & Inc-v3$_{ens3}$ & Inc-v3$_{ens4}$ & IncRes-v2$_{ens}$ & HGD & R\&D & NIPS-r3\\
         \cmidrule(r){1-1}
         \cmidrule(r){2-2}
         \cmidrule(r){3-11}
         \multirow{6}*{\shortstack{FGSM \\ \cite{goodfellow2014explaining}}} & normal RN18 & 47.3 & 40.7 & 33.8 & 33.0 & 34.6 & 23.1 & 26.2 & 24.9 & 26.2 \\
         & DSM(RN18) & \textbf{56.3} & \textbf{48.4} & \textbf{43.6} & \textbf{40.0} & \textbf{42.1} & \textbf{29.2} & \textbf{34.2} & \textbf{32.2} & \textbf{34.7} \\
         \cmidrule(r){2-11}
         & normal DN121 & 44.6 & 39.3 & 34.4 & 31.0 & 32.2 & 22.1 & 24.0 & 23.8 & 25.2 \\
         & DSM(DN121) & \textbf{54.1} & \textbf{47.9} & \textbf{43.0} & \textbf{40.0} & \textbf{39.9} & \textbf{29.3} & \textbf{33.5} & \textbf{31.1} & \textbf{32.9}  \\
         \cmidrule(r){2-11}
         & normal MNv2 & 42.4 & 34.2 & 28.2 & 26.8 & 28.3 & 17.9 & 18.4 & 19.2 & 22.3 \\
         & DSM(MNv2) & \textbf{46.0} & \textbf{40.0} & \textbf{32.7} & \textbf{30.2} & \textbf{30.8} & \textbf{20.9} & \textbf{21.5} & \textbf{22.1} & \textbf{24.7} \\
         \midrule
         \multirow{6}*{\shortstack{MI-FGSM\\ \cite{dong2018boosting}}} & normal RN18 & 62.0 & 54.2 & 43.7 & 39.2 & 39.0 & 26.2 & 35.2 & 28.7 & 31.7  \\
         & DSM(RN18) & \textbf{80.9} & \textbf{71.5} & \textbf{66.3} & \textbf{56.8} & \textbf{56.0} & \textbf{40.2} & \textbf{56.1} & \textbf{43.0} & \textbf{48.3} \\
         \cmidrule(r){2-11}
          & normal DN121 & 58.3 & 52.8 & 49.2 & 38.1 & 38.6 & 27.1 & 38.1 & 29.6 & 31.2  \\
          & DSM(DN121) & \textbf{79.7} & \textbf{77.4} & \textbf{70.4} & \textbf{55.1} & \textbf{51.9} & \textbf{40.7} & \textbf{56.9} & \textbf{41.8} & \textbf{46.3} \\
          \cmidrule(r){2-11}
          & normal MNv2 & 49.2 & 42.5 & 34.6 & 30.5 & 30.3 & 20.2 & 25.0 & 21.4 & 25.9 \\
          & DSM(MNv2) & \textbf{59.2} & \textbf{50.2} & \textbf{42.4} & \textbf{36.4} & \textbf{37.0} & \textbf{24.3} & \textbf{30.6} & \textbf{27.2} & \textbf{30.3} \\
         \midrule
         \multirow{6}*{\shortstack{TI-DI-MI-TG\\ \cite{he2022boosting}}} & normal RN18 & 81.7 & 75.3 & 66.4 & 58.9 & 58.8 & 42.8 & 55.8 & 49.5 & 54.4  \\
         & DSM(RN18) & \textbf{92.4} & \textbf{89.9} & \textbf{84.0} & \textbf{78.4} & \textbf{76.1} & \textbf{63.0} & \textbf{77.4} & \textbf{69.1} & \textbf{74.1} \\
         \cmidrule(r){2-11}
         & normal DN121 & 79.8 & 75.5 & 70.2 & 56.8 & 54.4 & 44.1 & 58.1 & 48.7 & 52.1  \\
         & DSM(DN121) & \textbf{92.6} & \textbf{92.1} & \textbf{89.6} & \textbf{77.9} & \textbf{75.5} & \textbf{63.0} & \textbf{81.7} & \textbf{69.2} & \textbf{74.4}  \\
         \cmidrule(r){2-11}
         & normal MNv2 & 72.2 & 65.3 & 59.5 & 49.4 & 49.6 & 34.4 & 43.9 & 38.7 & 44.8  \\
         & DSM(MNv2) & \textbf{82.4} & \textbf{75.3} & \textbf{70.0} & \textbf{59.7} & \textbf{59.0} & \textbf{43.4} & \textbf{54.9} & \textbf{47.2} & \textbf{54.9} \\
         \bottomrule
    \end{tabular}
\end{table*}

Then, we consider a more challenging dataset, ImageNet, and used nine publicly available models as victim models. These models have been widely used in previous work~\cite{dong2018boosting}. The first three of them are normally trained models, including Inception-v3 (Inc-v3)~\cite{szegedy2016rethinking}, Inception-v4 (Inc-v4), and Inception-ResNet-v2 (IncRes-v2)~\cite{szegedy2017inception}. The rest are robust models: Inc-v3$_{ens3}$, Inc-v3$_{ens4}$, and IncRes-v2$_{ens}$~\cite{tramer2018ensemble}, high-level representation guided denoiser (HGD)~\cite{liao2018defense}, input transformation through resizing and padding (R\&P)~\cite{xie2017mitigating}, and the rank-3 submission in NIPS2017 adversarial competition (NIPS-r3)\footnote{https://github.com/anlthms/nips-2017/tree/master/mmd}. To demonstrate the effectiveness of the proposed DSM across various adversarial example optimizers, we consider three optimizers, namely FGSM, MI-FGSM and TI-DI-MI-TG. The stronger TI-DI-MI-TG will be used as the default optimizer in the later section. We list the untargeted attack results in Table~\ref{tab:imagenet-un}, which shows that the proposed DSMs consistently outperform the normal surrogate models with same architecture. Notice that TI-DI-MI-TG represents a state-of-the-art method for generating adversarial examples without training a special surrogate model. Compared with it, using the three DSMs based on RN18, DN121 and MNV2 can improve the attack success rate by  10.7\%-21.6\%, 12.8\%-23.6\% and 8.5\%-11.0\%, respectively. 


\subsection{Ablation Studies}
\begin{table*}[h]
  \setlength{\abovecaptionskip}{0 cm}
    \centering
    \caption{The success rates (\%) of untargeted attacks using the DSMs with various teacher models on ImageNet-compatible dataset. 
    }
    \label{tab:imagenet-teacher}
    \begin{tabular}{@{}c@{~}c@{~}c@{~}c@{~}c@{~}c@{~}c@{~}c@{~}c@{~}c@{~}c@{}}
        \toprule
         Surrogate  model & Teacher model & Inc-v3 & Inc-v4 & IncRes-v2 & Inc-v3$_{ens3}$ & Inc-v3$_{ens4}$ & IncRes-v2$_{ens}$ & HGD & R\&D & NIPS-r3\\
         \cmidrule(r){1-1}
          \cmidrule(r){2-2}
           \cmidrule(r){3-11}
        \multirow{3}*{DSM(RN18)} & RN18 & 92.4 & 89.9 & 84.0 & 78.4 & 76.1 & 63.0 & 77.4 & 69.1 & 74.1 \\
         & DN121 & 90.5 & 87.4 & 80.5 & 70.3 & 67.9 & 55.4 & 70.3 & 60.1 & 64.8 \\
         & MNv2 & \textbf{95.0} & \textbf{91.3} & \textbf{87.6} & \textbf{83.4} & \textbf{80.2} & \textbf{66.9} & \textbf{81.5} & \textbf{71.3} & \textbf{77.6}  \\
         
         \midrule
         \multirow{3}*{DSM(DN121)} & RN18 & \textbf{97.8} & \textbf{95.9} & \textbf{94.5} & \textbf{89.7} & \textbf{88.9} & \textbf{80.9} & \textbf{91.0} & \textbf{85.4} & \textbf{87.6} \\
         & DN121 & 92.6 & 92.1 & 89.6 & 77.9 & 75.5 & 63.0 & 81.7 & 69.2 & 74.4 \\
          & MNv2 & 96.4 & 95.6 & 92.5 & 89.1 & 86.7 & 75.0 & 88.0 & 80.3 & 85.2\\
         \midrule
         \multirow{3}*{DSM(MNv2)} & RN18 & \textbf{89.5} & \textbf{86.7} & \textbf{80.7} & \textbf{71.3} & \textbf{67.9} & \textbf{55.6} & \textbf{70.2} & \textbf{60.6} & \textbf{67.0} \\
         & DN121 & 82.3 & 78.6 & 68.9 & 57.9 & 56.1 & 41.9 & 53.1 & 44.9 & 52.5 \\
         & MNv2 & 82.4 & 75.3 & 70.0 & 59.7 & 59.0 & 43.4 & 54.9 & 47.2 & 54.9 \\
         \bottomrule
    \end{tabular}
\end{table*} 

In this subsection, we first conducted experiments using teacher models with different architectures to investigate the effect of the teacher model and report the results in Table~\ref{tab:imagenet-teacher}. Notice that the results for the DSM sharing same architecture as the teacher model are the same as those in Table~\ref{tab:imagenet-un} for TI-DI-MI-TG. From Table~\ref{tab:imagenet-teacher} we see that using different teacher model may further improve the attack success rates. Comparing the results in Table~\ref{tab:imagenet-un}, we find out that DSMs can improve the attack success rates by \textbf{25.7\%}, \textbf{36.8\%} and \textbf{26.3\%} at most for the situations with RN18, DN121 and MNV2 based surrogate models, respectively. Although it is still an open problem that what teacher model is best for the adversarial transferability of DSM, just using the teacher model with the same architecture as DSM is a simple yet effective choice. 



\begin{figure}[h]
  \setlength{\abovecaptionskip}{0 cm}
  \centering
    \includegraphics[width=1.7in]{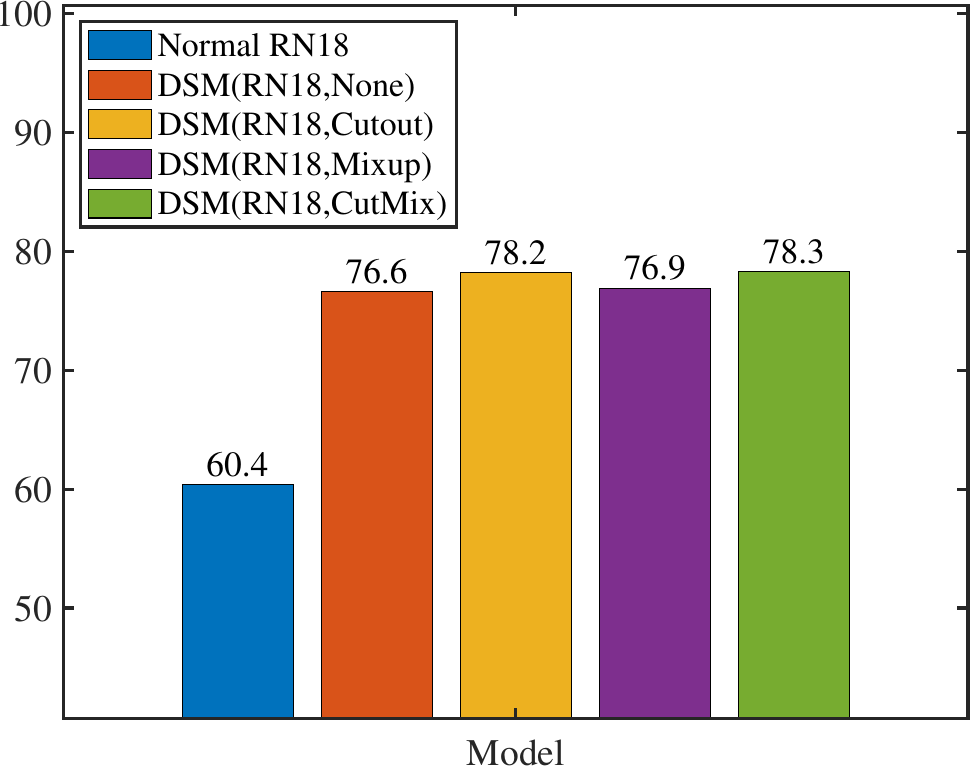}
    \caption{The average success rates (\%) of untargeted attack on ImageNet-compatible dataset against nine victim models when using different mixing augmentations to train a (dark) RN18.}
      \label{fig:exp_bar}
\end{figure}

We proceeded to investigate the impact of different mixing augmentations on adversarial transferability. Note that some preliminary results in this aspect were previously shown in Table~\ref{tab:cifar}. Here, we additionally consider two commonly used mixing augmentation skills, namely Cutout and Mixup. The RN18 is considered as the architecture of the surrogate model. The results are shown in Fig.~\ref{fig:exp_bar}, which shows that all mixing augmentation skills can improve the attack success rates. It is noteworthy that the improvement is smaller for ImageNet-compatible dataset than for CIFAR-10 (Table~\ref{tab:cifar}) due to the greater complexity of ImageNet. Consequently, the output of the teacher model does not degenerate to one-hot labels as depicted in Fig.~\ref{fig:pred}. Nevertheless, the use of mixing augmentation consistently improves adversarial transferability in all cases, with negligible additionally computational overhead.

Since the labeling strategy in DSM is different from the standard mixing augmentation skills, we also conduct experiments on a RN18 surrogate model trained with standard Cutout/Mixup/CutMix skills. Note that the only difference between a normal RN18 trained with standard mixing augmentation skill and a dark RN18 trained with the same mixing augmentation skill is the labeling strategy. The results of the RN18 models trained with these skills are presented in Table~\ref{tab:exp_bar_no_distill}. Surprisingly, we observed that the normal RN18 achieves better results than the models trained with mixing augmentation skills, indicating that such skills actually impairs the adversarial transferability of surrogate model when there is no teacher model to extract dark knowledge. This finding highlights the crucial role of extracted dark knowledge in boosting adversarial transferability.

\begin{table}[t]
  \setlength{\abovecaptionskip}{0 cm}
    \centering
    \caption{The success rates (\%) of untargeted attack on ImageNet-compatible dataset when using different strategies.}
    \label{tab:exp_bar_no_distill}
    \resizebox{0.5 \textwidth}{!}
    { 
    \begin{tabular}{@{}c@{~}c@{~}c@{~}c@{~}c@{~}c@{~}c@{~}c@{~}c@{~}c@{}}
        \toprule
         Surrogate  model & Inc-v3 & Inc-v4 & IncRes-v2 & Inc-v3$_{ens3}$ & Inc-v3$_{ens4}$ & IncRes-v2$_{ens}$ & HGD & R\&D & NIPS-r3\\
         \cmidrule(r){1-1}
         \cmidrule(r){2-10}
        Normal RN18 & \textbf{81.7} & \textbf{75.3} & \textbf{66.4} & \textbf{58.9} & \textbf{58.8} & \textbf{42.8} & \textbf{55.8} & \textbf{49.5} & \textbf{54.4}  \\
        RN18+Cutout & 79.5 & 73.5 & 65.8 & 58.4 & 55.7 & 41.9 & 55.0 & 45.9 & 50.3 \\
        RN18+Mixup & 77.8 & 72.0 & 62.6 & 53.7 & 50.5 & 36.6 & 47.5 & 39.1 & 42.0 \\
        RN18+CutMix & 74.0 & 67.0 & 57.4 & 51.7 & 50.6 & 37.1 & 46.5 & 40.0 & 43.7 \\
        \bottomrule
    \end{tabular}
    }
\end{table}


\subsection{Comparison with the Knowledge Distillation Based Method~\cite{cui2020substitute}}

\begin{figure}[h]
  \setlength{\abovecaptionskip}{0 cm}
  \centering
    \includegraphics[width=2.0in]{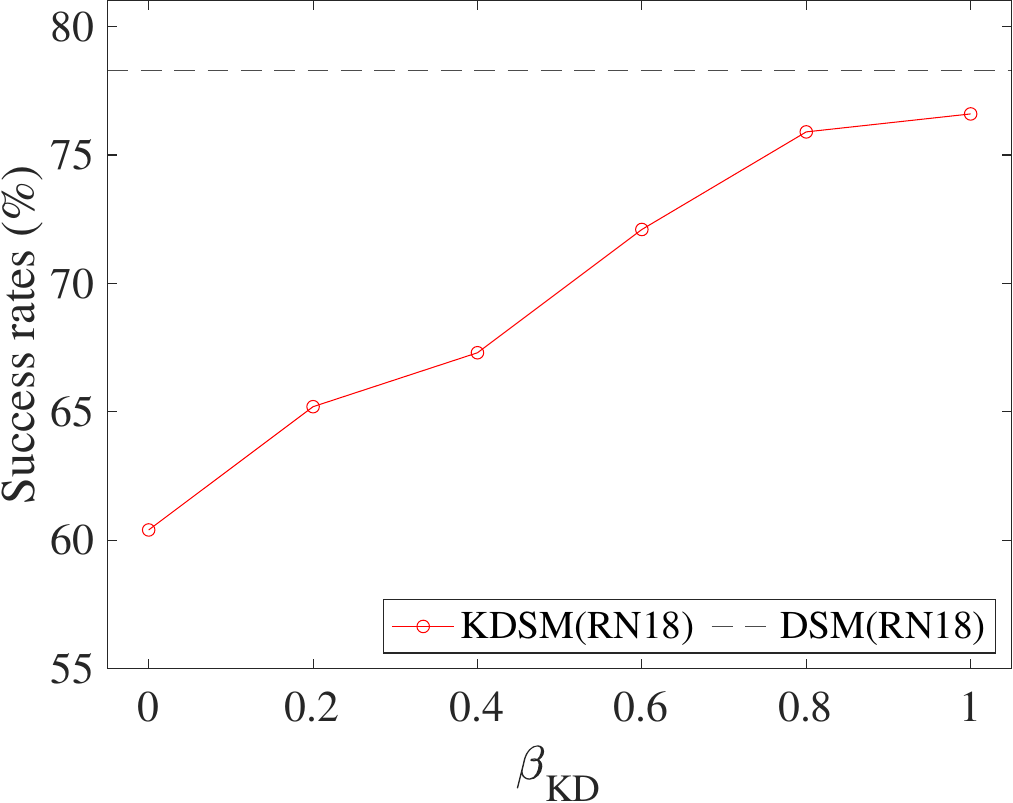}
  \caption{The average success rates (\%) of untargeted attack on ImageNet-compatible dataset against nine victim models with the KDSM~\cite{cui2020substitute} as the surrogate model, versus the value of $\beta_{KD}$ in~\eqref{eq:eq_KD}. The dash line indicates the result of DSM. 
  }
    \label{fig:beta}
\end{figure}

In this subsection, we compare the proposed DSM with the knowledge-distillation based method~\cite{cui2020substitute}, since it also uses knowledge distillation to train the surrogate model. We refer to the model trained in this way as KDSM (knowledge-distillation based surrogate model). If the number of teacher models $M=1$, the difference between our DSM and KDSM is that the latter uses one-hot labels to interpolate with dark knowledge (see~\eqref{eq:eq_KD}), while the former additionally uses the mixing augmentations to enhance dark knowledge. 
Using normal RN18 as the teacher model to train a surrogate model in same architecture, the results of KDSM and our DSM are shown in Fig.~\ref{fig:beta}. 
When $\beta_{KD}=0$, the KDSM degenerates to a normally trained model, and when $\beta_{KD}=1$ the KDSM is equivalent to our DSM without mixing augmentation. Fig.~\ref{fig:beta} shows that the attack success rate increases as $\beta_{KD}$ increases, and it reaches the maximum at $\beta_{KD}=1$. This means the DSM is better than KDSM, and it is unnecessary and inadvisable  to interpolate with the one-hot labels. 
Furthermore, as Fig.~\ref{fig:exp_bar} once demonstrated, the dash line in Fig.~\ref{fig:beta} shows better results for DSMs with mixing augmentation. The effectiveness of mixing augmentation to enhance the dark knowledge is also observed in other dataset. For example, the results presented in Table~\ref{tab:cifar} demonstrate that mixing augmentation improves the attack success rates by up to \textbf{27.0\%} on CIFAR-10, while there is almost no improvement without using mixing augmentation. Additionally, Section IV.E will verify the effectiveness of mixing augmentation in attacking face verification models.

\subsection{The Results of Targeted Attack}

\begin{table}[b!]
    \setlength{\abovecaptionskip}{0pt}
    \setlength{\belowcaptionskip}{0pt}

    \centering
    \caption{The success rates (\%) of targeted attacks on ImageNet-compatible dataset.}
    \label{tab:imagenet-tar}
    \resizebox{0.5 \textwidth}{!}
    { 
    \begin{tabular}{@{}c@{~}c@{~}c@{~}c@{~}c@{~}c@{~}c@{~}c@{~}c@{~}c@{}}
        \toprule
         Surrogate  model & Inc-v3 & Inc-v4 & IncRes-v2 & Inc-v3$_{ens3}$ & Inc-v3$_{ens4}$ & IncRes-v2$_{ens}$ & HGD & R\&D & NIPS-r3\\
         \cmidrule(r){1-1}
         \cmidrule(r){2-10}
        SR-RN18~\cite{springer2021a} & 46.2 & 41.6 & 49.0 & 38.3 & 40.4 & 34.2 & 49.5 & 39.0 & 41.0 \\
         DSM(SR-RN18) & \textbf{63.1} & \textbf{59.7} & \textbf{65.5} & \textbf{48.9} & \textbf{49.0} & \textbf{41.1} & \textbf{62.2} & \textbf{48.2} & \textbf{51.7} \\
        \bottomrule
    \end{tabular}
    }
\end{table}
We have shown that the proposed DSM performs very well on untargeted attack. While for the much more difficult task of targeted attack, a state-of-the-art work is the method with slightly robust surrogate model~\cite{springer2021a} despite that it costs large computational time for generating adversarial examples online. Below, we will show that it can be further improved with the proposed DSM while almost not inducing extra time cost. The combination of DSM and the method with slightly robust surrogate model~\cite{springer2021a} can make a favorable success rate of targeted attack.
In~\cite{springer2021a}, it is demonstrated that 
by using a slightly robust model trained with small-magnitude adversarial examples as the surrogate model, the state-of-the-art success rates on targeted attack are achieved. 
Specifically, using $N$ rounds of iteration to generate adversarial examples would make the training time to be $N+1$ times that of standard training, where $N$ is about 10 as suggested by~\cite{pang2020bag}. We use the slightly robust ResNet18 model (SR-RN18), which was trained with maximum perturbation of 0.1 (the recommend valued in~\cite{springer2021a}), as the teacher model to train a DSM denoted by DSM(SR-RN18), taking about the same time as standard training. We conduct experiments of targeted attacks and report the results in Table~\ref{tab:imagenet-tar}, which shows that the proposed DSM can be naturally combined with~\cite{springer2021a} and it again remarkably improves the success rates of black-box attack by 6.9\%-18.1\%.
\subsection{Application to Attacking Face Verification Model}
DNN models for face verification have been widely deployed in many safety-critical scenarios like mobile unlocking. To show the versatility of the proposed method, we present the experimental results on attacking face verification models in this subsection. A face verification model is used to judge whether a pair of facial images belong to the same identity. It is built based on a classifier trained on a dataset of facial images to separate images of different identities. Given a pair of facial images, a pair of embedding features are extracted by the classifier, i.e. the outputs of the penultimate layer of the model. Then, the cosine similarity between them is calculated for judging whether they belong to the same identity.

The dodging attack and impersonate attack are two kinds of attack to face verification model. Given a pair of facial images $x$ and $x_r$ belonging to the same identity, dodging attack aims to generate an adversarial example $x^{adv}$ which is similar to $x$ but be recognized as a different identity from $x_r$. On the contrary, impersonate attack aims to generate an adversarial example $x^{adv}$ which is similar to $x$ but be recognized as the same as $x_r$ if $x$ and $x_r$ do not belong to the same identity. We conduct experiments on the standard LFW~\cite{huang2008labeled} protocol, which means we select both 3000 pairs of images for dodging attack and impersonate attack. The IResNet50 (IR50)~\cite{he2016identity} is chosen as the surrogate model and four publicly available face verification models as the victim models are considered, including FaceNet, SphereFace, CosFace and ArcFace since they have different architectures and are considered in prior works on attacking face verification models~\cite{yang2020delving}.

We train an IR50 classifier on CASIA-WebFace~\cite{yi2014learning} following previous work~\cite{yang2020delving}, and use it as a teacher model to train the dark surrogate model. 
\begin{table}
    \setlength{\abovecaptionskip}{0pt}
    \setlength{\belowcaptionskip}{0pt}
  \centering
  \caption{The success rates (\%) of the dodging/impersonate attacks to face verification models on LFW dataset.}
    \label{tab:face}
    \resizebox{0.5  \textwidth}{!}
    {
    \begin{tabular}{@{}c@{~}c@{~}c@{~}c@{~}c@{~}c@{~}c@{~}c@{~}c@{}}
    \toprule
         \multirow{2}*{Surrogate model}& \multicolumn{4}{c}{Dodging attack} & \multicolumn{4}{c}{Impersonate attack} \\
         \cmidrule(r){2-5}
      \cmidrule(r){6-9}
& FaceNet & SphereFace & CosFace & ArcFace & FaceNet & SphereFace & CosFace & ArcFace\\
            \midrule
         IR50 & 79.2 & 95.6 & 93.2 & 77.4 & 45.4 & 84.5 & 76.3 & 60.6 \\
        DSM(IR50,None) & 86.2 & 97.7 & 96.2 & 84.2 & 53.6 & 88.6 & 82.0 & 69.9 \\
        DSM(IR50,CutMix) & \textbf{92.5} & \textbf{99.4} & \textbf{98.8} & \textbf{90.3} & \textbf{63.0} & \textbf{93.8} & \textbf{87.2} & \textbf{76.8} \\
         \bottomrule
    \end{tabular}
    }
\end{table}
We conduct dodging/impersonate attack experiments on them with $\epsilon=8$ and list the results in Table~\ref{tab:face}, which shows that adversarial transferability can be remarkably improved through using dark knowledge, and can be further improved by introducing CutMix. Specifically, the proposed DSM can improve the success rates of dodging attack and impersonate attack by $\textbf{12.9\%}$ and $\textbf{16.2\%}$ respectively, when the ArcFace~\cite{deng2019arcface} is used as the victim model.

\section{Conclusions}
In this paper, we propose a method to train the surrogate model for transfer-based adversarial attack on image classification, which boosts the adversarial transferability of surrogate models. The trained surrogate model is named dark surrogate model (DSM). The proposed method includes two key components: using a teacher model to generate dark knowledge (soft label) for training the surrogate model, and using the mixing augmentation skills to enhance the dark knowledge of training data. The effectiveness of the proposed method is validated by extensive experiments and the comparisons with counterparts. Besides, we show that the proposed method can be extended to other transfer-based attack scenarios that contain dark knowledge, like face verification.
\section{Acknowledgment}
This work was supported by the National Key Research and Development Plan of China (2020AAA0103502), and National Key Research and Development Project of China (No. 2021ZD0110502).
\bibliographystyle{IEEEtran}
\bibliography{IEEE}

\end{document}